\documentclass[10pt,twocolumn,letterpaper]{article}

\usepackage{cvpr}
\usepackage{graphicx}
\usepackage{amsmath}
\usepackage{amssymb}
\usepackage{times}
\usepackage{epsfig}
\usepackage{multirow}
\usepackage{booktabs}
\usepackage{subcaption}
\usepackage{array}
\usepackage{caption}
\usepackage{arydshln}
\usepackage{arydshln}
\usepackage{float}
\usepackage{paralist}
\usepackage{xcolor}
\usepackage{wrapfig}

\usepackage{pifont}
\newcommand{\cmark}{\ding{51}}%
\newcommand{\xmark}{\ding{55}}%
\usepackage[export]{adjustbox}
\usepackage{enumitem}

\usepackage[pagebackref=true,breaklinks=true,letterpaper=true,colorlinks,bookmarks=false]{hyperref}

\begin{document}

\title{Weakly Supervised Face and Whole Body Recognition in Turbulent Environments}

\author{Kshitij Nikhal\\
University of Nebraska-Lincoln\\
{\tt\small knikhal2@huskers.unl.edu}
\and
Benjamin S. Riggan\\
University of Nebraska-Lincoln\\
{\tt\small briggan2@unl.edu}
}

\maketitle
\thispagestyle{empty}

\begin{abstract}
   Face and person recognition have recently achieved remarkable success under challenging scenarios, such as off-pose and cross-spectrum matching. However, long-range recognition systems are often hindered by atmospheric turbulence, leading to spatially and temporally varying distortions in the image. Current solutions rely on generative models to reconstruct a turbulent-free image, but often preserve photo-realism instead of discriminative features that are essential for recognition. This can be attributed to the lack of large-scale datasets of turbulent and pristine paired images, necessary for optimal reconstruction. To address this issue, we propose a new weakly supervised framework that employs a parameter-efficient self-attention module to generate domain agnostic representations, aligning turbulent and pristine images into a common subspace. Additionally, we introduce a new tilt map estimator that predicts geometric distortions observed in turbulent images. This estimate is used to re-rank gallery matches, resulting in up to 13.86\% improvement in rank-1 accuracy. Our method does not require synthesizing turbulent-free images or ground-truth paired images, and requires significantly fewer annotated samples, enabling more practical and rapid utility of increasingly large datasets. 
   We analyze our framework using two datasets---Long-Range Face Identification Dataset (LRFID) and BRIAR Government Collection 1 (BGC1)---achieving enhanced discriminability under varying turbulence and standoff distance.  
\end{abstract}
\vspace{-0.4cm}

\begin{figure}[ht]
    \centering
    \fbox{\includegraphics[width=\columnwidth]{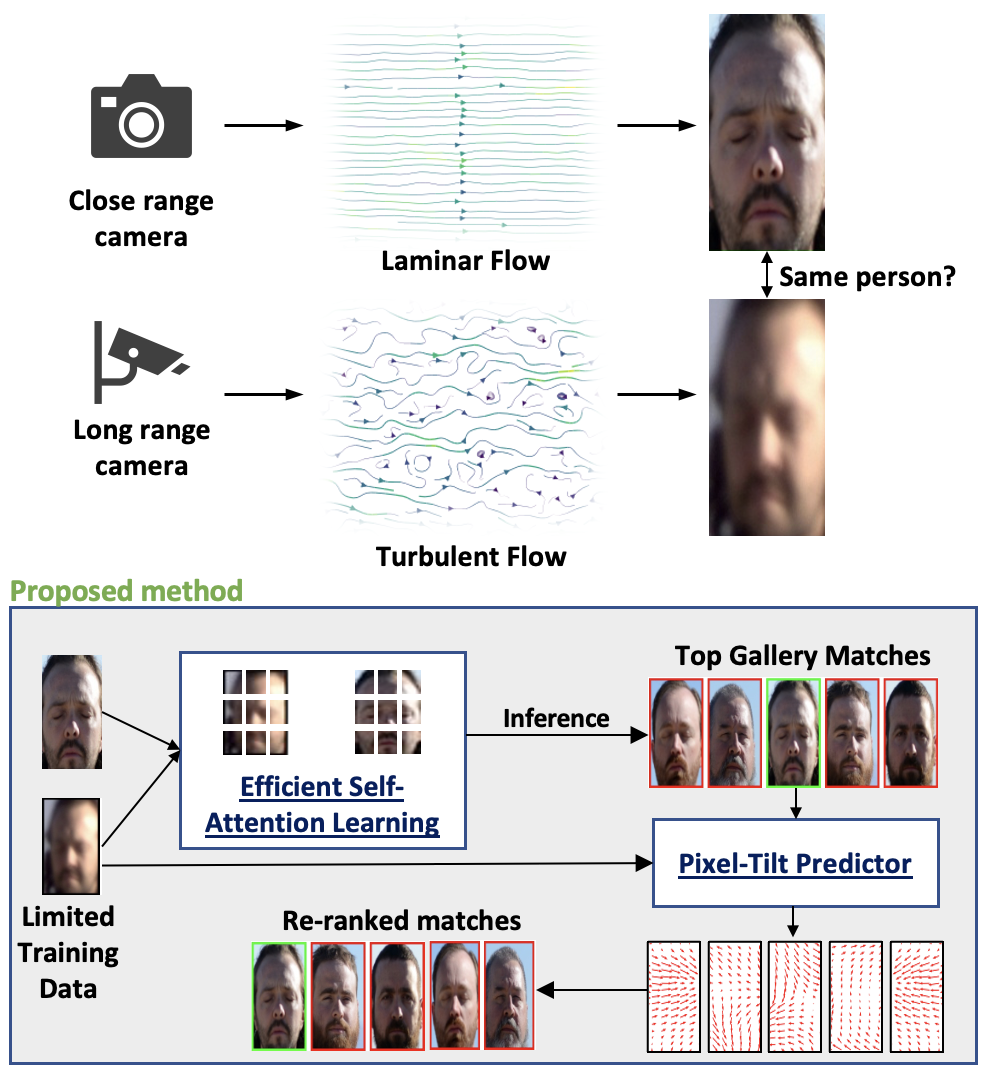}}
    \caption{Overview of our proposed method. }
    \label{fig:overview}
\end{figure}

\section{Introduction}
Supervised face and person (whole-body) recognition methods have continued to demonstrate enhanced discriminability in increasingly challenging scenarios, such as profile-to-frontal~\cite{sengupta2016frontal, mostofa2022pose}, infrared-to-visible~\cite{fondje2020cross, chen2005ir}, and low-to-high resolution matching~\cite{zou2011very, massoli2020cross}.  These improvements can be attributed to several factors, including the increased availability of large-scale curated (annotated) datasets, parallelism afforded by better graphics processing unit (GPU) technology and associated application programming interfaces (APIs), and complexity of machine learning models. However, our ability to effectively and efficiently annotate and exploit increasing amounts of data is quickly becoming impractical due to time and resource constraints.  This is especially true for imagery acquired for multi-domain operations, such as operations security (OPSEC), force protection, border patrol, criminal justice, and counter-terrorism, which exhibit degraded quality due to unconstrained conditions such as extended standoff distances ($100$--$500m$). 
A common stochastic effect observed when capturing imagery from long-range is atmospheric turbulence, which is caused by perturbations of the air particles of different densities. Digital imaging through turbulence results in severe loss of image quality due to random tilt---spatially varying geometric distortions caused by local changes in the index of refraction---and optical blur.  Recently, restoration of turbulent imagery has been studied~\cite{Mao2022SingleFA, nair2023ddpm, 9506614, mao2020image, zhang2022imaging}, but is difficult to perform in practice and at scale due to the lack of paired ground-truth data. 
While these restoration methods show some promising ability to recover coarse information from turbulent imagery, several potential limitations or concerns arise, including 
\emph{(1) necessary recovery of fine-grained details to achieve optimal recognition,
(2) requirements for complex generative networks (exceeding 10 million parameters) to restore images,
and (3) demands for exponentially large and precisely annotated datasets.} 

Instead of bridging the nonlinear effects of turbulence by attempting to simultaneously learn inverse effects of both tilt and blur, we bridge the turbulence gap by applying tilt maps---estimated from long-range query (probe) images---to gallery images.  This procedure geometrically distorts (or warps) non-turbulent, close-range gallery images in a way that is consistent with long-range imagery, resulting in a direct alignment of discriminative regions of face and whole-body images. This avoids the need to synthesize turbulent-free images, which enables more reliable matching between query and gallery images for recognition purposes. Moreover, since tilt maps are more efficiently and reliably estimated compared to reconstructing high-quality images, this unique approach is better suited for weak supervision---optimization using significantly fewer annotated (labeled) samples---and compact embedding networks that map long- and close-range images to a common latent subspace.  

Therefore, the primary outcome from this paper is a new \emph{weakly supervised} framework for face and whole-body recognition (Figure~\ref{fig:overview}), composed of the following individual components and their contributions:
\begin{itemize}[noitemsep,topsep=-0.5pt]
    \item a \emph{parameter-efficient self-attention module} that uses domain agnostic intermediate representations to align turbulent and pristine images into a common subspace,
    \item a new \emph{tilt map estimator} that predicts the geometric distortion (pixel shifts) observed from query images,   
    \item a novel \emph{re-ranking approach} that applies estimated tilt maps to gallery imagery to improve recognition.
\end{itemize}
 We analyzed our framework using the Long-Range Face Identification Dataset (LRFID)~\cite{lrfid} and BRIAR\footnote{Biometric Recognition \& Identification at Altitude and Range} Government Collection 1 (BGC1)~\cite{briar}, which are challenging datasets for long-range recognition.  While there are ethical concerns, responsible informed consent and data storage protocols have been followed. 
We achieve enhanced discriminability under varying turbulence and standoff distance on both datasets, and establish a new benchmark on BGC1 whole body dataset using a small fraction (e.g., less than 20\% supervision) of the labeled data.
\begin{figure*}[ht]
    \centering
    \includegraphics[width=1.8\columnwidth]{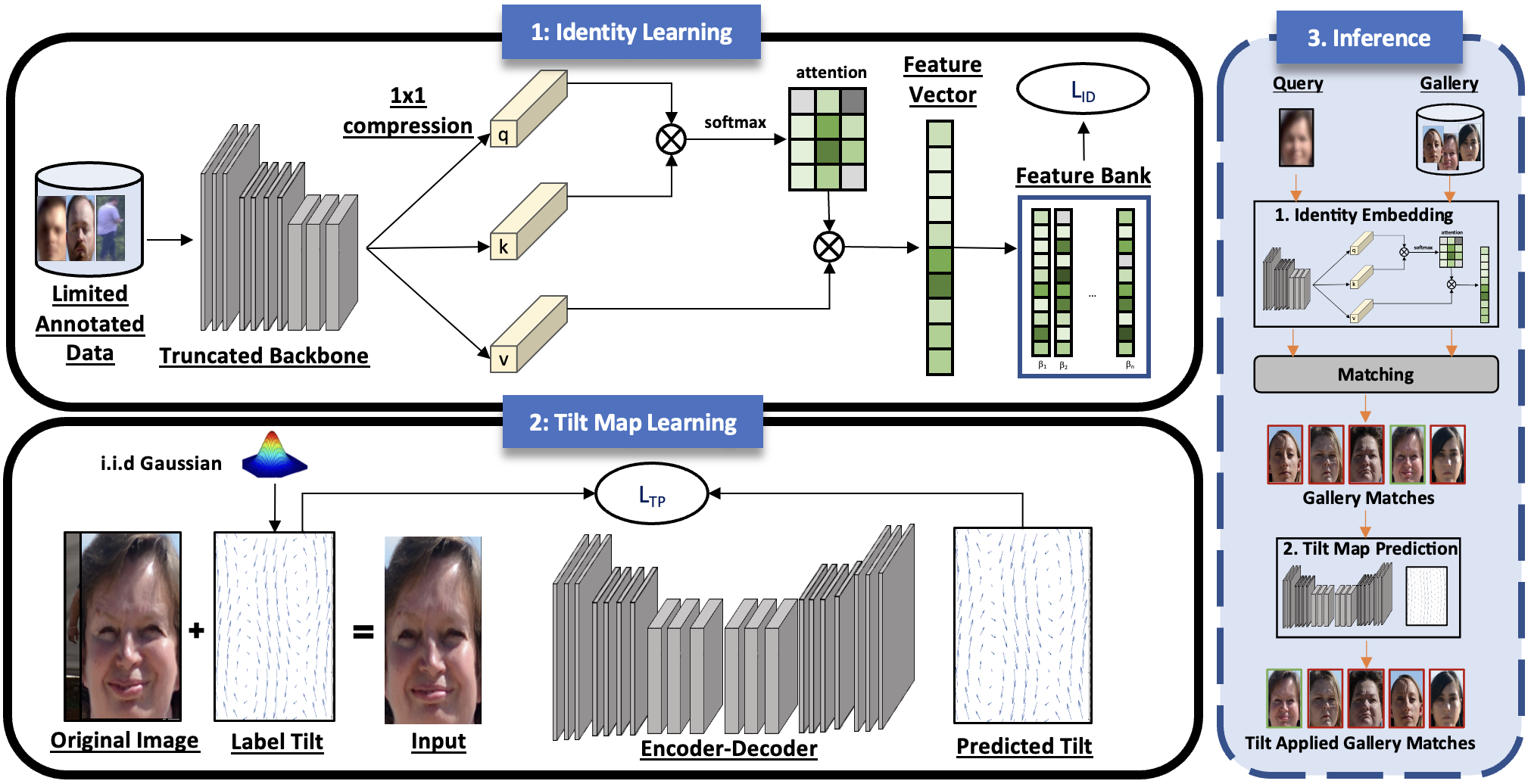}
    \caption{Our proposed method has two training phases: (1) Identity Learning where intermediate features are transformed using a self-attention module to generate domain agnostic representations and (2) Tilt Map Learning where the geometric distortion of turbulent images is learned by augmenting pristine images with random spatial fields. During inference, the identity model generates embeddings to compare query and gallery images. The top-5 matches are further re-ranked by applying the learned tilt from the turbulent query image.  }
    \label{fig:methodology}
\end{figure*}

     
     

\section{Relevant Literature}
\subsection{Learning with Limited Supervision:}
Learning with limited supervision has gained attention due to the high cost and complexity of annotating biometric data. Semi-supervised learning often uses contrastive learning to ensure same representation is produced on original and augmented images~\cite{yang2022survey}. 
A newer paradigm called meta-learning has been studied on the few-shot learning problem, to adapt to newer tasks using prior experiences~\cite{finn2017model, nichol2018reptile}. However, the lack of cross-domain labeled data in long-range settings make it challenging to  transfer knowledge. 
SimCLR~\cite{simclr} shows that composition of multiple data augmentation operations and very large batch-sizes ( $>2048$) benefits contrastive learning, yielding discriminative representations. FixMatch~\cite{fixmatch} combines consistency regularization and pseudo-labeling on weak and strong augmentations of the same input. 
While this has achieved success on some tasks~\cite{simclr, fixmatch}, strong augmentation is insufficient to bridge large domain gaps and in most cases, degrades performance. Unsupervised methods~\cite{lin2019bottom, nikhal2021unsupervised, nikhal2022multi} use instance-based pseudo-labeling and agglomerative clustering to progressively merge similar samples together to optimize the representation. However, the large gap between pristine and turbulent domains has difficulty to cluster inter-domain samples together. 
This work proposes a method to transfer intermediate features, that has coarse but general features, to adapt to newer tasks with limited data.

\subsection{Turbulence Mitigation:}
Most methods attempt to eliminate turbulence and treat it as an image restoration task. 
Early methods took a physics-based approach to restore images, with optical adaptive techniques~\cite{Pearson, roggemann1996imaging} and local information fusion~\cite{Vorontsov}.
In recent years, deep learning based methods have been proposed in the literature.
In~\cite{yasarla2020learning}, a Monte Carlo simulation is used to model blur and degradation to restore turbulence.~\cite{7536179} reconstructs a high-quality image from observed images through low-rank decomposition. In~\cite{lau2019restoration}, the deformation field between image frames is estimated and warped using quasiconformal maps. AT-DDPM~\cite{nair2023ddpm} uses Diffusion Probabilistic Models to iteratively denoise turbulent images. AtFaceGAN~\cite{lau2020atfacegan} disentangles the blur and deformation operations and generates an image with sharp details. Physics-based models~\cite{Mao2022SingleFA, Mao_2021_ICCV} capture the long-range dynamics of turbulence effects. However, the synthetically restored image creates another domain gap (ideally reduced) that needs to be aligned. 
In~\cite{wes2022}, the effect of feature shift under turbulence is studied, showing that face recognition models misinterpret turbulence as salient features. In this work, we guide the generation of salient features using pixel-level self attention and directly align inter-domain features. 

\section{Methodology}
Our proposed framework is shown in Figure~\ref{fig:methodology}. 
First, we  generate a strong domain-invariant representation for turbulent and pristine images. Next, we train a tilt-predictor to predict the geometric distortion of the turbulent image. Finally, we apply this distortion on the gallery images to re-rank the top matches to boost rank-1 performance of the recognition system. The parameter efficient design of the self-attention module makes our framework suitable for weak supervision, requiring significantly fewer annotated samples. Moreover, our method does not depend on corresponding pairs of turbulent and pristine images.

\subsection{Preliminaries}
We denote gallery (or pristine) images as $X_{G} = \{x^g_1, x^g_2, \dots, x^g_n\}$ and query (or turbulent) images as $X_{Q} = \{x^q_1, x^q_2, \dots, x^q_n\}$ and the corresponding identity labels as $Y_{G} = \{y^g_1, y^g_2, \dots, y^g_n\}$ and $Y_{Q} = \{y^q_1, y^q_2, \dots, y^q_n\}$. Note that the images (gallery or query) from the training and evaluation set are disjoint, and no optimization is done on the evaluation set. To start, we randomly initialize a memory bank $M = \{\beta_1, \beta_2, \dots, \beta_{nc}\}$ where $nc$ is the number of known classes (or identities). 
We address closed-set person and face recognition, where the input query image is a subject captured from long-range, and we aim to match it with the close-range images (e.g., driver's license or passport photos) of the same subject from the gallery/database. Here, closed-set implies the assumption that subjects in query images have a mated image (or template) in the gallery. 

\setlength{\abovedisplayskip}{2pt}
\setlength{\belowdisplayskip}{2pt}

\subsection{Weakly Supervised Identification}
\label{sec:id_learning}
To bridge the domain gap between pristine $X_G$ and turbulent $X_Q$ images with weak supervision, we extract intermediate features from a convolutional backbone network:
\begin{equation}
    F_{bb} = backbone(x^d_i)~\text{for}~i \dots n,
\end{equation}
where $d \in \{g, q\}$. It is important to emphasize that $X_G$ and $X_Q$ here denote the images from the training set. 
We empirically determine $stage~3$ of the ResNet50~\cite{he2016deep} network to extract intermediate features, as they are sufficiently (and initially) discriminative but still retain generic features. 
These intermediate features are then fed to three (query, key, and value) $1 \times 1$ convolutional transforms:
\begin{equation}
F^s = tanh(Conv^s_{1024->512}(F_{bb}))~~s \in \{q, k, v\}
\end{equation}
that compress feature map from 1024 to 512 channels while retaining spatial dimensions.

The query and key are matrix multiplied together and the softmax operation is applied to calculate the self-attention:
\begin{equation}
    F_{attn} = softmax(F^{q} \cdot F^k.T).
\end{equation}
Lastly, the attention is multiplied by the value feature to generate the final identity representation. 
\begin{equation}
    F_{final} = F^v \cdot F_{attn}
\end{equation}
While some equations are similar to the original self-attention model proposed in ~\cite{vaswani2017attention}, the notable differences are (a) the use of a $1 \times 1$ convolutional transform that forces representations to retain salient but concise features by compression, (b) the use of self-attention as a transformation (or projection head) rather than inserting in between layers and (c) the use of $tanh$ to introduce non-linearity.  

The final representation is optimized using an adaptable memory bank $M$ that stores the centers---mean feature representations 
The probability of image $x_i$ belonging to it's cluster center is: 

\begin{equation}
    P(B_i, x_i) = \frac{exp(M^{T}_{\beta_i}~ f(x_i)) / \tau}{\sum_{j \neq i} exp(M^{T}_{\beta_j}~f(x_i)) / \tau}
\end{equation}
where $f(x_i)$ is the mapping from the input image to the final representation and $\tau$ is the temperature parameter of the distribution set to 0.1 as in ~\cite{nikhal2022multi}.
Then, the objective function for identity learning is the negative log likelihood:

\begin{equation}
    L_{ID} = - \sum^n_{i=1} \log (P(\beta_i|x_i)),
\end{equation}
that minimizes intra-class variance and maximizes inter-class separability. 

\subsection{Tilt Map Predictor}
While the identity learner bridges the domain gap between turbulent and pristine images to a degree, we notice relatively higher rank-5  and low rank-1 recognition scores. This means that the network was able to retrieve the correct identity in the top-5 matches. To this end, we propose to integrate
the distortions produced by isoplanatic turbulence, i.e., the pixel shifts are spatially varying with a constant blur. Ignoring the spatial variance of the blur and simulating the tilt is widely used in image reconstruction literature ~\cite{anantrasirichai2013atmospheric, lau2019restoration, repasi2011computer, leonard2012simulation}. 
To do this, we aim to predict the pixel shift (or tilt) of the given turbulent image on a coarse-level. The hypothesis is that applying the predicted tilt on the gallery images will boost facial recognition performance. 

\begin{figure}
    \centering
    \includegraphics[width=0.7\columnwidth]{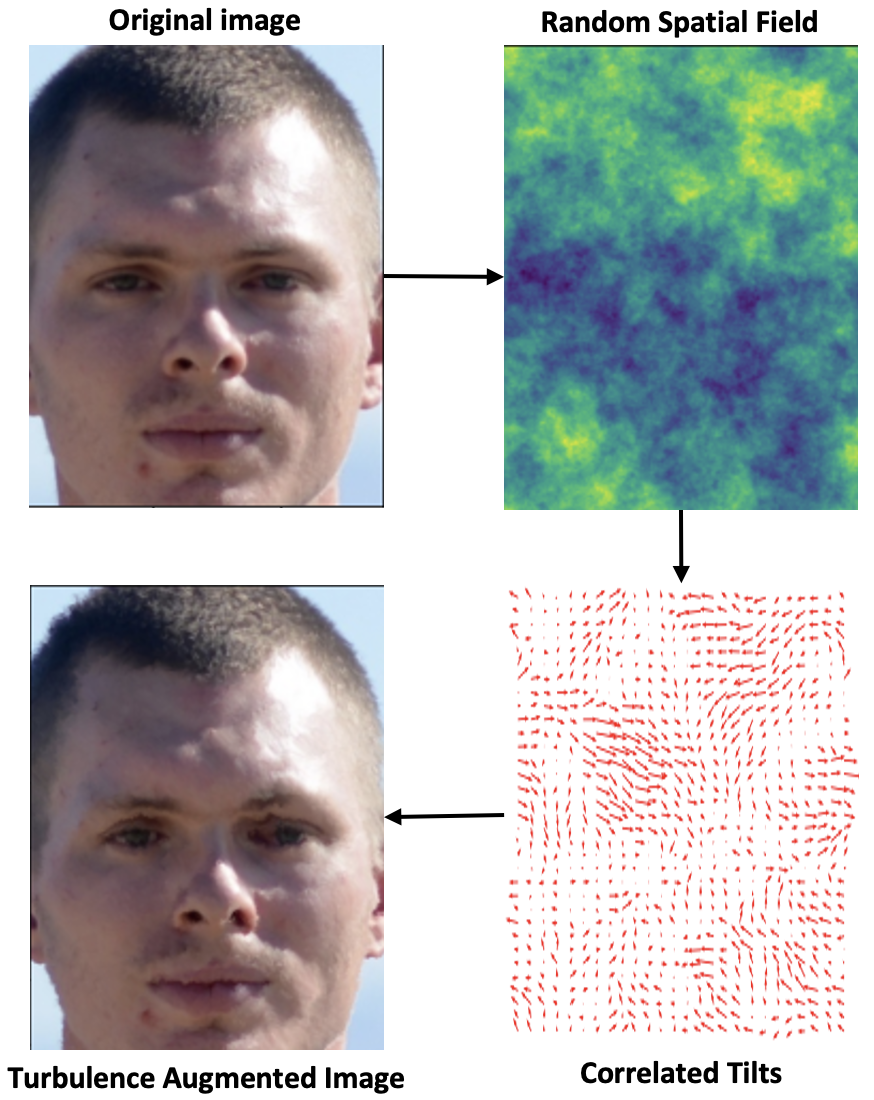}
    \caption{Geometric distortion is introduced to pristine (gallery) images, warping specific facial features like eyes, nose, mouth, jaw line, and head shape found in turbulent images. Subject has consented to image publication.  }
    \label{fig:distortion}
\end{figure}

We train an encoder-decoder architecture with a ResNet50 encoder pretrained on ImageNet  to predict this effect. As no ground-truth pixel shifts are available for the dataset, we use pristine images from the training set augmented by a spatially correlated random Gaussian field to mimic this effect. This field is generated by (i) generating a complex Gaussian noise with normal distribution $N\sim\mathcal{CN}(\mu=0, \sigma=1)$, (ii) multiplying the noise and the power law spectrum  $N \times P(k)$, and (iii) taking the inverse Fourier transform of this field, i.e., $Re\{\mathcal{F}^{-1}(N \times P(k))\}$. The resulting real component is a Gaussian random field with spatial correlations resulting from a scale-free power law spectrum:
\begin{equation}
    P(k) = 1/k^{\alpha}
\end{equation} where $k$ is the amplitude of the power law and $\alpha$ is the power law exponent. A value of $\alpha$ close to 1 produces a smooth field whereas a value of 0 produces a rough field. 
Figure~\ref{fig:distortion} shows the result on one input image. The supplementary material includes a video that shows the distortion results. Using the power law spectrum to generate random fields is common in astrophysics~\cite{pen1997generating} and the Kolmogorov spectrum $\alpha=5/3$ is often used to model many turbulent flows~\cite{kolmogorov1941local} and hence we set it to this number. We vary the correlation length of the spatial field---the largest eddying motion ranges of the turbulent flow---for each image to make the network robust to varying ranges of turbulence in the image. The decoder is trained to predict the spatial field that distorted the image.

Mathematically, we apply a spatial pixel distortion $\Psi$ (resulting in a pixel tilt) on the image with field $f$ varying correlation length. The network $d(\cdot)$ is optimized by: 
\begin{equation}
    L_{TP} = MSE(d(\Psi(x_i^g;f_i)), f_i)
\end{equation}
to minimize the difference between the real and predicted spatial field. 
After convergence, we use this model to predict the pixel shift on turbulent images. 

\subsection{Re-Ranking using Tilt Prediction}
For a given turbulent query image $q_k$ from the testing set, we predict the distortion field $f_{k} = d(q_k)$ and apply this field to all the images (from the testing set) in the gallery:
\begin{equation}
    \hat{g_i} = \Psi(x_i^g, f_k)~\forall~i~\text{in}~X_G
\end{equation}
Since we achieve high rank-5 performance (using the identity embedding in Section~\ref{sec:id_learning}) and the gallery size is much larger than 5, we restrict the use of this technique to the top-5 gallery matches for a given query. This keeps the re-ranking method efficient and does not degrade performance for low quality distortion predictions. We also consider Gaussian blur and conclude that applying the tilt leads to a greater improvement in rank-1 performance compared to using a combination of blur and tilt, or solely applying blur, as seen in Table~\ref{tab:tiltblur}. 
\begin{wraptable}{l}{4cm}
\caption{Combining tilt and blur}
\label{tab:tiltblur}

\centering
\begin{tabular}[h]{|c|c|c|c|}
    \hline
         Tilt & Blur & Rank-1 \\
    \hline
         \xmark & \xmark & 64.93 \\
         \cmark & \xmark & \textbf{70.12} \\
         \xmark & \cmark & 66.23 \\
         \cmark & \cmark & 68.83 \\
    \hline
    \end{tabular}
\end{wraptable}
    Visually, we observe that image warping based on tilt inherently introduces a degree of blur, rendering additional blur inefficient in boosting performance. \\
\\



\begin{table}[h]
    \caption{Whole body recognition  performance on the BGC1 dataset. The bracket specifies the dataset the model is pretrained on. \textbf{Bold} denotes best performance. }
    \centering
    \resizebox{1.0\columnwidth}{!}{
    
    \begin{tabular}{lllllll}
    \cmidrule{1-7}
     Range & Method & Venue & Rank-1 & Rank-3 &  Rank-5 & mAP \\
    

    \cmidrule(lr){1-7}
    

    {\multirow{8}{*}{100m}}
    & \vline~PCB(BGC1)~\cite{sun2018beyond} & ECCV18 & 11.25 & 21.25 & 35.00 & 23.97 \\
    & \vline~PCB(Market1501)~\cite{sun2018beyond} & ECCV18 & 20.62 & 38.75 & 48.75 & 35.17 \\
    & \vline~OSNet(Market1501)~\cite{zhou2021learning} & ICCV19 & 26.88 & 40.00 & 49.38 & 38.19 \\
    & \vline~OSNet(BGC1)~\cite{zhou2021learning} & ICCV19 & 27.50 & 39.38 & 56.88 & 40.71 \\
    & \vline~BPBNet(BGC1)~\cite{somers2023body} & WACV23 & 17.50 & 30.00 & 40.00 & 30.71 \\
    & \vline~BPBNet(Market1501)~\cite{somers2023body} & WACV23 & 16.88 & 32.50 & 40.00 & 30.50 \\
    & \vline~OURS & IJCB23 & 31.25 & 62.50 & 79.37 & 50.78\\
    & \vline~OURS+TP & IJCB23 & \textbf{33.75} & \textbf{63.12} & \textbf{79.37} & \textbf{50.78} \\
    \cmidrule(lr){1-7}
    
    {\multirow{8}{*}{200m}}
    & \vline~PCB(BGC1)~\cite{sun2018beyond}  & ECCV18  & 12.87 & 24.56 & 35.09 & 25.18 \\
    & \vline~PCB(Market1501)~\cite{sun2018beyond}  & ECCV18  & 26.90 & 46.20 & 54.97 & 40.48 \\
    & \vline~OSNet(Market1501)~\cite{zhou2021learning}  & ICCV19 & 26.32 & 43.86 & 56.14 & 40.45 \\
    & \vline~OSNet(BGC1)~\cite{zhou2021learning} & ICCV19 & 14.04 & 34.50 & 47.37 & 30.97 \\
    & \vline~BPBNet(BGC1)~\cite{somers2023body}  & WACV23  & 8.77 & 25.73 & 39.77 & 23.93 \\
    & \vline~BPBNet(Market1501)~\cite{somers2023body}  & WACV23 & 25.73 & 43.86 & 50.88 & 38.88 \\
    
    & \vline~OURS & IJCB23 & 30.99 & 52.05 & 62.57 & \textbf{45.57} \\
    & \vline~OURS+TP & IJCB23 &  \textbf{33.91} & \textbf{52.94} & \textbf{62.94} & 45.09 \\
    
    \cmidrule(lr){1-7}

    {\multirow{8}{*}{400m}}
    & \vline~PCB(BGC1)~\cite{sun2018beyond}  & ECCV18 & 9.86 & 25.35 & 33.80 & 23.26 \\
    & \vline~PCB(Market1501)~\cite{sun2018beyond}  & ECCV18 & 23.94 & 39.44 & 47.18 & 36.67 \\
    & \vline~OSNet(Market1501)~\cite{zhou2021learning} & ICCV19 & 21.13 & 33.80 & 42.25 & 32.98 \\
    & \vline~OSNet(BGC1)~\cite{zhou2021learning} & ICCV19 & 16.20 & 38.03 & 49.30 & 32.44 \\
    & \vline~BPBNet(BGC1)~\cite{somers2023body} & WACV23 & 14.08 & 27.46 & 38.73 & 26.01 \\
    & \vline~BPBNet(Market1501)~\cite{somers2023body} & WACV23 & 19.72 & 30.28 & 39.44 & 30.02 \\
    
    & \vline~OURS & IJCB23 & 43.66 & \textbf{72.54} & 77.46 & \textbf{59.47} \\
    & \vline~OURS+TP & IJCB23 &  \textbf{45.77} & 71.12 & \textbf{77.47} & 59.46 \\

    \cmidrule(lr){1-7}
    
    \multirow{8}{*}{500m} 
    & \vline~PCB(BGC1)~\cite{sun2018beyond} & ECCV18 & 8.76 & 18.25 & 27.74 & 19.55  \\
    & \vline~PCB(Market1501)~\cite{sun2018beyond} & ECCV18 & 20.44 & 39.42 & 48.18 & 34.50\\
    & \vline~OSNet(Market1501)~\cite{zhou2021learning} & ICCV19 & 18.98 & 37.23 & 40.88 & 31.89 \\
    & \vline~OSNet(BGC1)~\cite{zhou2021learning} & ICCV19 & 10.22 & 24.09 & 38.69 & 24.56 \\
    & \vline~BPBNet(BGC1)~\cite{somers2023body} & WACV23 & 8.76 & 24.09 & 37.23 & 22.60 \\
    & \vline~BPBNet(Market1501)~\cite{somers2023body} & WACV23 & 16.06 & 33.58 & 42.34 & 30.17 \\
    & \vline~OURS & IJCB23 & 33.58 & \textbf{56.93} & 71.53 & 49.72\\
    & \vline~OURS+TP & IJCB23 & \textbf{47.44} & 55.47 &\textbf{71.53} & \textbf{49.72}\\

    \cmidrule(lr){1-7}

    \end{tabular} 
    }
    \label{tab:briar}
\end{table}

\begin{figure*}[h]
    \centering
    \includegraphics[width=2.1\columnwidth]{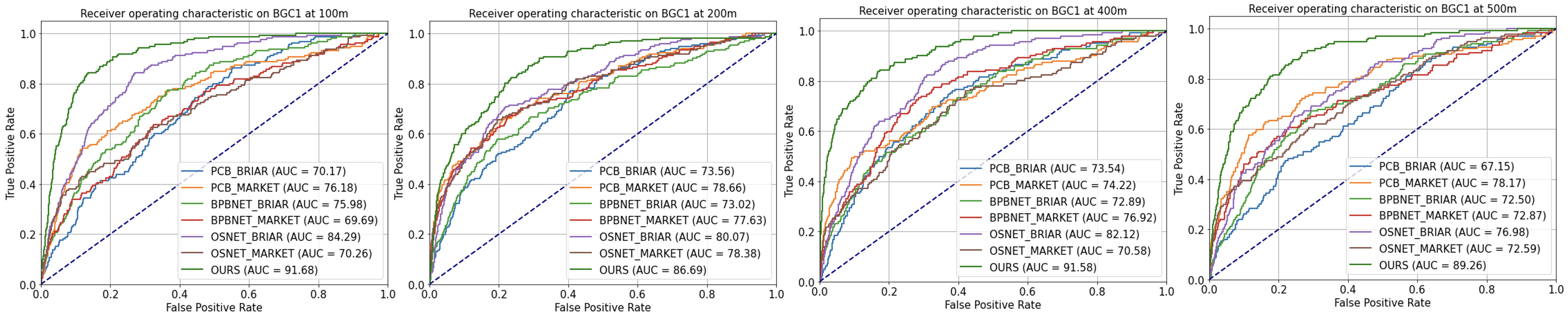}
    \caption{ROC Curves on the BGC1 dataset where our method (\textcolor{green}{green curve}) significantly outperforms other methods in verification scores.}
    \label{fig:roc_curves}
\end{figure*}

\begin{table*}[h]
    \caption{Face Recognition performance on LRFID and BGC1 face dataset. * denotes performance reported in~\cite{nair2023ddpm}}. 
    \centering
    \resizebox{1.4\columnwidth}{!}{
    
    \begin{tabular}{lllllllllll}

    \cline{3-10}
    \multicolumn{1}{c}{} & \multicolumn{1}{c}{} \vline & \multicolumn{4}{c}{LRFID} \vline & \multicolumn{4}{c}{BGC1} \vline \\
    \hline
 Method & Venue & Rank-1 & Rank-3 & Rank-5 & mAP & Rank-1 & Rank-3 & Rank-5 & mAP \\
    \hline
        
     ArcFace~\cite{deng2019arcface} & CVPR19 & 40.26 & 62.34 & 71.43 & 53.86 &  30.00 & 43.33 & 50.00 & 40.49 \\
     ATFaceGAN*~\cite{lau2020atfacegan} & FG20 & 47.50 & 65.80 & 82.30 & - & 22.00 & 38.00 & 50.00 & -  \\
     GFPGAN*~\cite{Wang_2021_CVPR} & CVPR21 & 57.30 & 79.20 & 85.30 & - & 26.00 & 58.00 & 60.00 & - \\
     
     MPR-NET*~\cite{zamir2021multi} & CVPR21 & 34.10 & 64.60 & 74.40 & - & 24.00 & 46.00 & 64.00 & - \\
     AT-NET*~\cite{9506614} & ICIP21 & 36.50 & 64.60 & 74.40 & - & 14.00 & 28.00 & 38.00 & - \\
     TurbNet~\cite{Mao2022SingleFA} & ECCV22 & 44.16 & 63.64 & 67.53 & 56.57 & 28.00 & 39.33 & 49.00 & 38.03  \\
     LTTGAN*~\cite{mei2023ltt} & JSTSP23 & 58.50 & 81.70 & 85.30 & - & 20.00 & 54.00 & 62.00 & -\\
     AT-DDPM*~\cite{nair2023ddpm} & WACV23 & 62.20 & 81.70 & \textbf{87.80} & - & 32.00 & \textbf{56.00} & \textbf{66.00} & - \\
     AT-DDPM~\cite{nair2023ddpm} & WACV23 & 31.17 & 45.45 & 57.14 & 32.78 & 21.00 & 28.00 & 40.00 & 32.48\\
     OURS & IJCB23 & 64.93 & 79.22 & 85.07 & 73.99 & 32.00 & 49.00 & 62.00 & 46.21\\
     OURS+TP & IJCB23 & \textbf{70.12} & \textbf{81.80} & 85.07 &\textbf{73.99} & \textbf{33.00} & 49.00 & 62.00 & \textbf{46.21}\\
    
    \hline
\end{tabular} 
    }
\label{tab:lrfid}
\end{table*}

\section{Experiments}
\textbf{Datasets:}
The LRFID~\cite{lrfid} was collected by the US Army C5ISR division to study facial recognition under extreme atmospheric effects. It contains indoor and outdoor enrollment (close-range) videos that are free of turbulence, and outdoor videos captured at 350m of 100 identities. 

The BGC1~\cite{briar} dataset comprises of still images and videos that are unconstrained and variable quality from close-range, 100m, 200m, 400m, and 500m distance ranges for 150 identities. Unlike LRFID, both face and whole-body (WB) images are acquired for purposes of human identification.  With multiple standoff distances, this dataset helps benchmark robustness of methods across varying levels of turbulence and resolution. 
We sample one frame from each video for each identity at a distance range for the testing protocol. 
We present one of the first analysis of whole body intra-set matching on this dataset. In addition, we also compare with existing methods for facial recognition. 

In this work, we adopt a weakly supervised approach by sampling less than 20\% of the identities for training and 80\% of the identities (not belonging to the training set) for testing.  
We follow the same testing protocol as ~\cite{nair2023ddpm} to be able to benchmark and demonstrate clear and fair comparison over previous methods. Both datasets have obtained approvals from the IRB (Institutional Review Board) and can be requested by contacting the respective authors. Code available at \url{https://git.unl.edu/ece-unl-images-lab/recognition-in-turbulent-environments}

\begin{figure*}[ht]
    \centering
    \includegraphics[width=1.8\columnwidth]{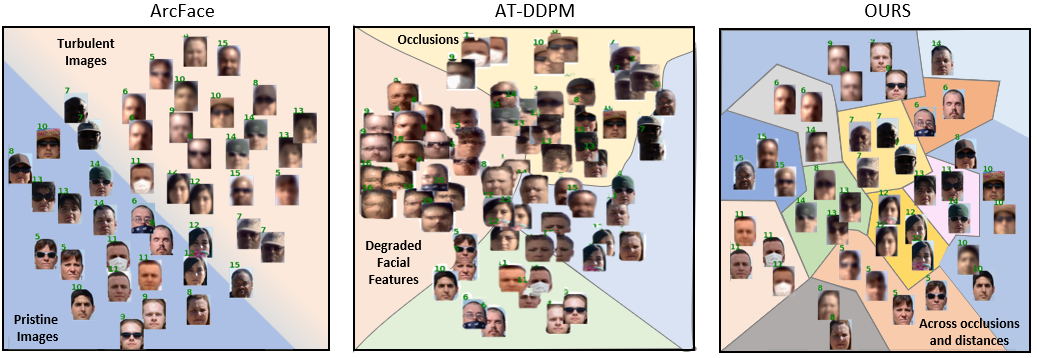}
    \caption{t-SNE plot showing ArcFace representations clusters turbulent and pristine images, AT-DDPM representations does not have clear inter and intra-class separation, while ours correctly clusters identities together across occlusions and environments. Subjects appearing in this figure consent to use their image in publications. }
    \label{fig:tsne}
\end{figure*}

\begin{figure*}[h]
    \centering
    \includegraphics[width=1.7\columnwidth]{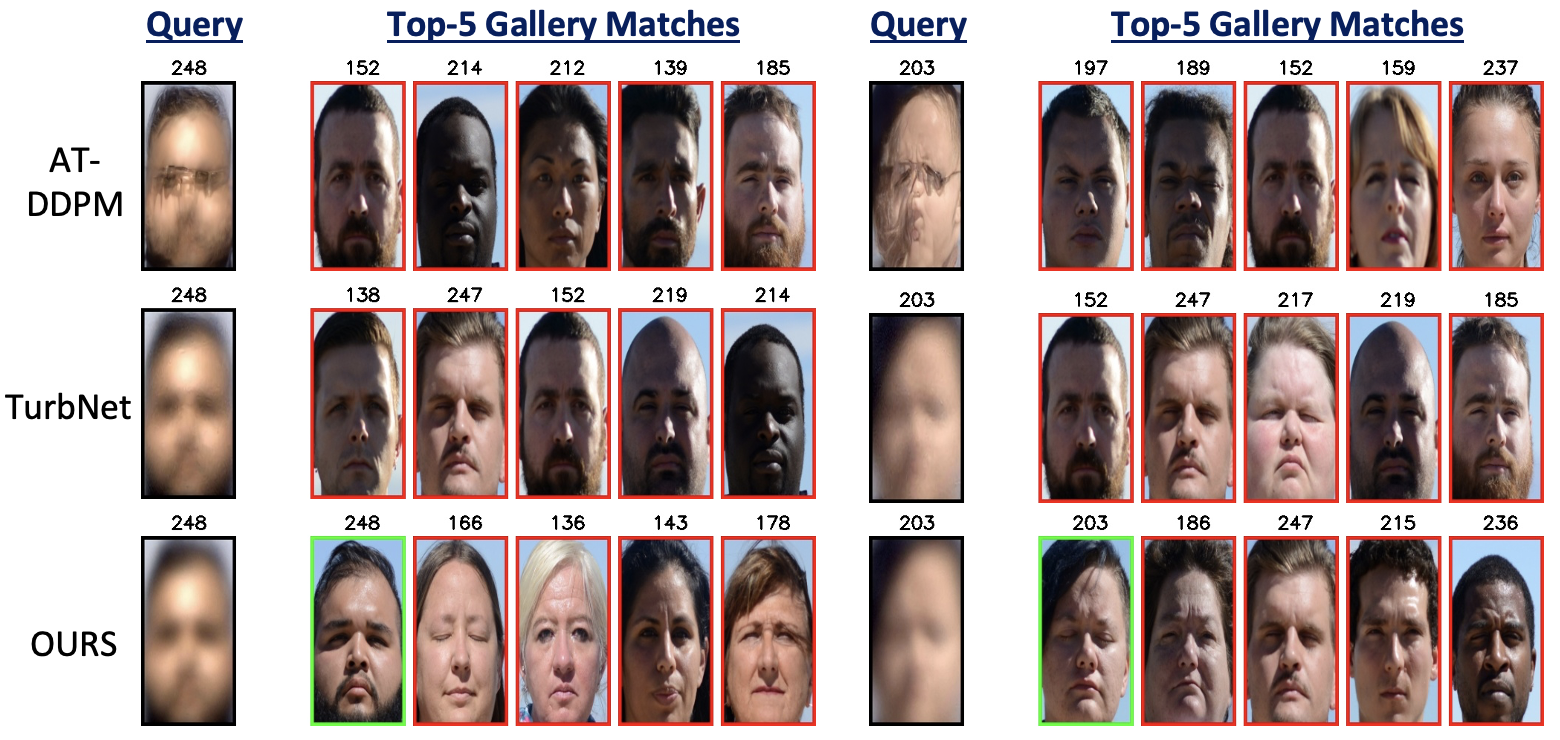}
    \caption{AT-DDPM and TurbNet visually improves quality of  images in the LRFID dataset, but suffer from degradation of salient facial features essential for discriminability. Our method does not synthesize new images but instead focuses on bridging the domain gap, resulting in improved discriminability. Subjects in results consent to image publication. }
    \label{fig:lrfid_ranking}
\end{figure*}

\subsection{Quantitative Results}
We present results using verification metrics, including the Receiver Operating Characteristics (ROC) curve---True Positive Rate (TPR) as a function of False Positive Rate (FPR)---and the Area under the Curve (AUC)---overall separability (across all operating thresholds) between genuine and imposter match scores. Additionally, we measure rank-$k$ accuracy, representing the number of correct retrievals in the highest $k$ matches, and the mean average precision (mAP), representing the average measure of separability of correct matches for each subject. 

Table~\ref{tab:briar} shows the results on whole body recognition on the BGC1 dataset across various distance ranges. We compare with seminal works in supervised and domain adaptable person re-identification, specifically PCB~\cite{sun2018beyond}, BPBNet~\cite{somers2023body}, and OSNet~\cite{zhou2021learning}. In low turbulence setting (100m and 200m), our method achieves an improvement of 3.75\% and 4.09\% improvement in rank-1 performance over recent methods. In addition, our tilt predictor further improves rank-1 performance by $\sim$2\%. Moreover, in high turbulence setting (400m and 500m), our method significantly outperforms previous methods by 19.72\% and 10.14\%. The tilt prediction (TP) also boosts performance by 2.11\% and 13.85\% in high turbulence setting. The relatively lower influence of tilt prediction in the 400m range may be attributed to the usage of different cameras, diminishing the impact of turbulence captured in the image. The results also indicate a noticeable improvement in mAP scores compared to recent methods across distances, demonstrating increased level of separability among correct predictions. 
Figure~\ref{fig:roc_curves} shows the results of verification scores across methods. Our method is reliable in the task of verification, with an AUC score improvement of 7.39, 6.62, 9.46, and 11.08 at the 100m, 200m, 400m, and 500m distances, respectively. 

Table~\ref{tab:lrfid} shows the facial recognition performance on the LRFID and BGC1 datasets. As previous methods are reported on random sampling of frames, we compare with reported scores (denoted by $*$) in ~\cite{nair2023ddpm} and also test the methods on our sampling. For LRFID, we notice an improvement of 29.86\% over state-of-the-art (SOTA) facial recognition ArcFace~\cite{deng2019arcface} used in the image reconstruction methods. TurbNet~\cite{Mao2022SingleFA} improves performance of ArcFace by around 4\%, and AT-DDPM's~\cite{nair2023ddpm} performance is impacted by degradation of some facial features as seen in Figure~\ref{fig:tsne}. Our method outperforms reported AT-DDPM score by 7.92\% in rank-1 scores. Furthermore, the improved separability of our method is demonstrated by the higher mAP scores achieved by our proposed technique. 
TP enhances rank-1 performance by 5.19\%, indicating that our re-ranking approach consistently improves performance across various datasets and distance ranges. Our performance on the BGC1 dataset is comparable to the reported scores of AT-DDPM since the dataset contains very low-resolution images with varying poses and motion blur. We do not address face pose correction in this study.

\begin{figure*}[h]
    \centering
    \includegraphics[width=1.6\columnwidth]{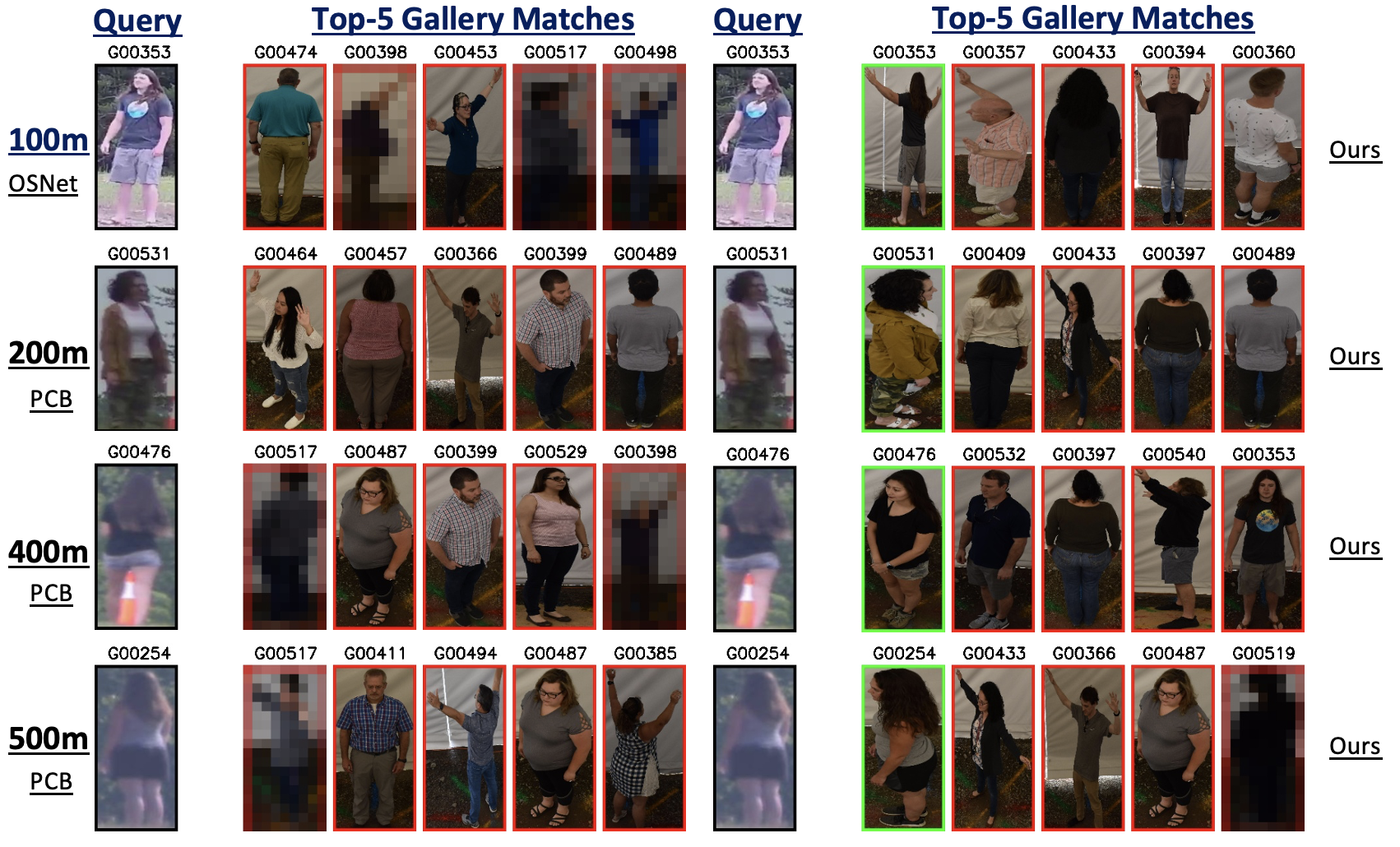}
    \caption{Comparison of BGC1 ranking results between the best-performing method (left) for that distance range and our method (right). Our model is robust to various poses and quality (row 1, 2, and 4) and occlusions (row 3). G00398, G00517, G00498, and G00519 did not consent to image publication and are pixelated for privacy. Others consented for image use in publications. }
    \label{fig:briar_ranking}
\end{figure*}
\subsection{Qualitative Results}
Figure~\ref{fig:tsne} shows the separability of features on the LRFID dataset using t-SNE~\cite{van2008visualizing}. ArcFace, intended for high resolution facial recognition, clusters pristine and turbulent images, resulting in subpar performance. AT-DDPM clusters occluded images (such as mask, hat, and sunglasses), while low quality are similarly grouped. In contrast, our method has better separability among identities and correctly clusters the same subject across occlusions and distance ranges. 

In Figure~\ref{fig:lrfid_ranking}, we see the ranking performance on the LRFID dataset compared to SOTA image reconstruction methods: AT-DDPM~\cite{nair2023ddpm} and TurbNet~\cite{Mao2022SingleFA}. Although AT-DDPM generates sharper images, the high turbulence results in artifacts in salient facial regions like the eyes and nose, leading to decreased performance compared to TurbNet. TurbNet uses physics-based method for reconstruction that produces fewer artifacts, but does not reconstruct salient regions that are used by off-the-shelf recognition systems such as ArcFace~\cite{deng2019arcface}. Our approach does not focus on  reconstruction, but instead directly overcomes the domain gap between turbulent and pristine images and retrieves the correct identity, as seen in row 3 of the figure. 

Similarly, Figure~\ref{fig:briar_ranking} shows the ranking results on the BGC1 dataset compared to the best-performing method on that dataset range. Our approach shows robustness to different poses and occlusions, varying turbulence, and is able to accurately retrieve the identity.

\subsection{Ablation Studies}
\noindent\textbf{Transformation Heads:}
In Table~\ref{tab:projection_heads}, we present a comparison of different transformation heads. CBAM~\cite{woo2018cbam} and Triplet Attention~\cite{misra2021rotate} both suffer from high parameterization and large kernel fields resulting in low performance. FCANet~\cite{qin2021fcanet} utilizes a fixed basis for data representation with minimal learning, resulting in slightly better performance due to the reduced risk of over-fitting. LKA~\cite{guo2022visual} models long-range pixel correlations that is efficient by using depth wise dilated convolutions, and achieves a rank-1 performance of 21.90\%. SAGAN~\cite{zhang2019self} uses $1 \times 1$ transformations similar to our method, but uses high degree of compression and number of layers resulting in over-fitting. 
Compared to the original self attention, we achieve an improvement of 4.38\% and 3.78\% in rank-1 and mAP scores, clearly showing the improvement of our modified self-attention transformation head. 

\begin{table}[h]
    \centering
     \resizebox{0.8\columnwidth}{!}{
    \begin{tabular}{|c|c|c|c|c|}
    \hline
         Method & Rank-1 & Rank-3 & Rank-5 & mAP\\
    \hline
         CBAM~\cite{woo2018cbam} & 13.87 & 22.63 & 29.20 & 23.07 \\
         FcaNet~\cite{qin2021fcanet} & 15.33 & 31.39 & 51.09 & 30.54 \\
         LKA~\cite{guo2022visual} & 21.90 & 38.69 & 49.64 & 35.33 \\
         Triplet Attention~\cite{misra2021rotate} & 11.68 & 17.52 & 27.01 & 20.09 \\
         SAGAN~\cite{zhang2019self} & 16.06 & 24.09 & 30.66 & 24.51\\
         AttentionConv~\cite{ramachandran2019stand} & 29.20 & 51.82 & 66.42 & 45.94 \\
         Ours & \textbf{33.58} & \textbf{56.93} & \textbf{71.53} & \textbf{49.72} \\
    \hline
    \end{tabular}
    }
    \caption{Comparison of transformation heads on BGC1 at 500m.}
    \label{tab:projection_heads}
\end{table}

\begin{figure}[h]
    \centering
    \includegraphics[width=0.7\columnwidth]{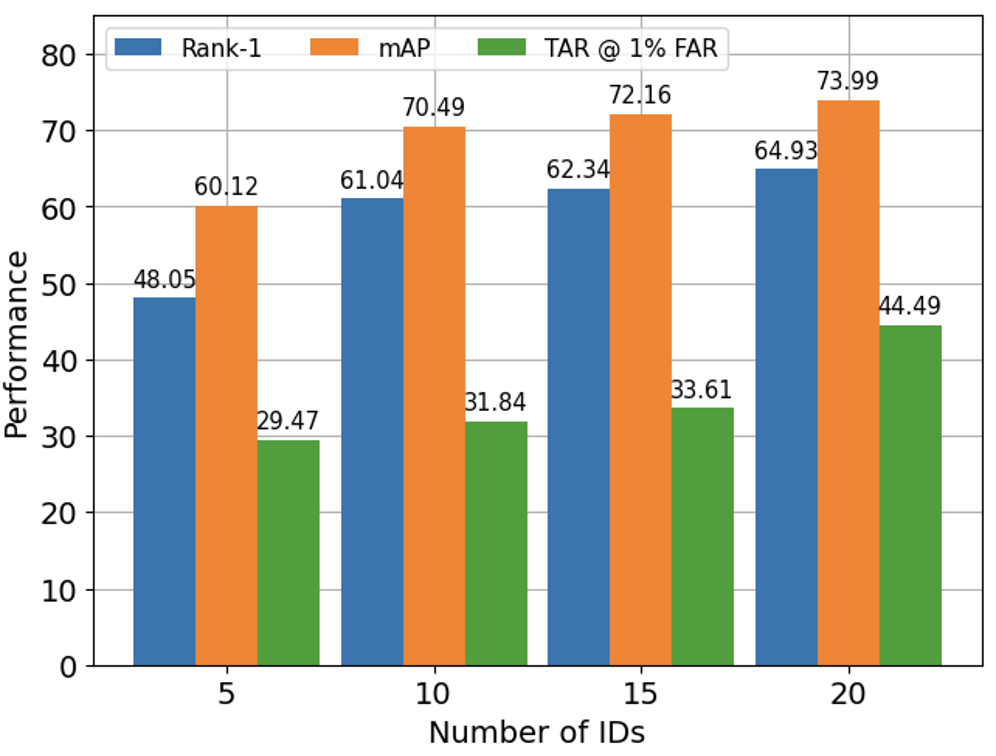}
    \caption{Progressive learning performance with only 5 identities.}
    \label{fig:progressive_learning}
\end{figure}

\noindent\textbf{Degree of Supervision and Progressive Learning: }
Figure~\ref{fig:progressive_learning} shows rank-1, mAP and TAR @ 1\% FAR performance on various levels of supervision. Using only 5 identities, we are able to attain a rank-1 performance of 48.05\%, mAP score of 60.12\%, and TAR @ 1\% FAR of 29.47\%. However, we notice a substantial gain of 12.99\% and 10.37\% in rank-1 and mAP scores, respectively, when the number of identities were increased to 10. With increasing level of supervision, we noticed that both rank-1 and mAP gradually increased while TAR showed a significant improvement. This indicates that higher levels of supervision result in an increased level of separability between positives and negatives. 
This trend continues with higher levels of supervision.

\begin{figure}[h]
    \centering
    \includegraphics[width=0.6\columnwidth]{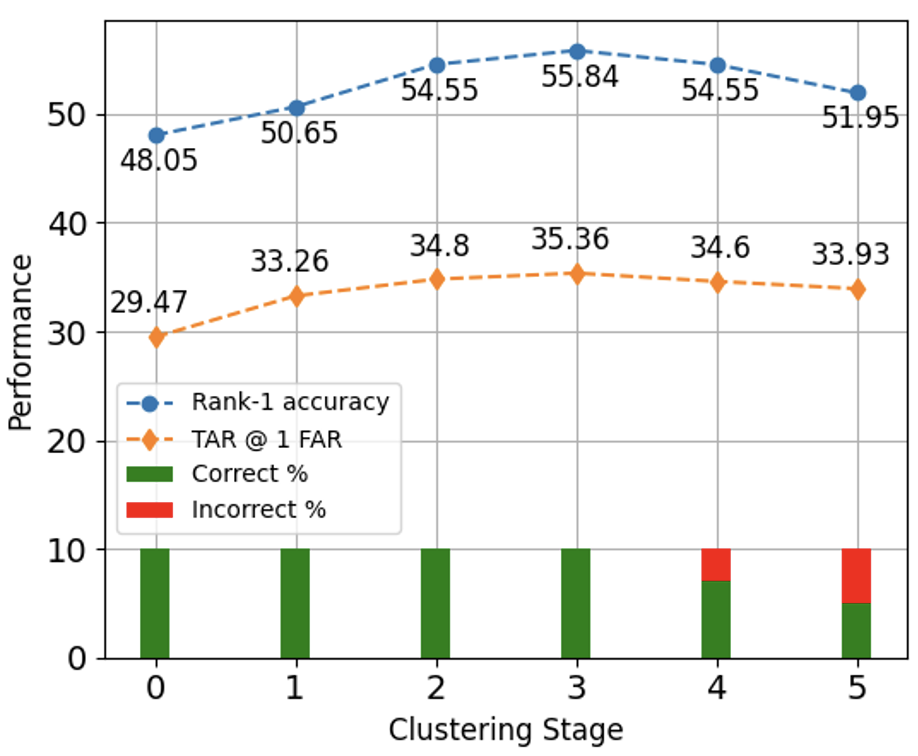}
    \caption{Rank and verification performance with varying supervision on the LRFID dataset. Results consistently improve with minimal supervision until incorrect matches are clustered together. }
    \label{fig:varying_supervision}
\end{figure}

Figure~\ref{fig:varying_supervision} shows results of incorporating progressive learning in our method by iteratively predicting on the unlabeled train set and incorporating that in the training phase. We start with 5 identities as the baseline and iteratively clustered around 100 frames, which improved performance for 4 stages of training. However, at stage 4, incorrect identities were matched together, resulting in a drop in performance. Although we do not address progressive learning in this work, future work can incorporate smart clustering~\cite{lin2019bottom, nikhal2022multi, hu2022divide} and robustness to noisy labels~\cite{song2022learning} to consistently improve performance.  



\vspace{-0.5cm}


\section{Conclusion}
Long-range and turbulent conditions significantly degrade the quality of face and whole-body recognition performance. Despite promising trends in developing restoration models and toward increasing complexity and supervision, we evaluated an alternative weakly supervised framework. Our framework learns domain-agnostic representations for matching turbulent and pristine imagery by leveraging a parameter-efficient self-attention transformation head with few annotated examples. Our new tilt-map estimator helps boost rank-1 performance by predicting the geometric distortion from query images to re-rank the gallery matches. Our framework is generalizable to both face and whole body recognition, and set new benchmarks across multiple datasets with varying levels of turbulence. 
Our research is committed to maintaining rigorous ethical and privacy standards, and responsible use, while aiming to maximize benefits to individuals and society. 
\section{Acknowledgement}
This research is based upon work supported in part by the
Office of the Director of National Intelligence (ODNI), Intelligence Advanced Research Projects Activity (IARPA),
via 2022-21102100002. The views and conclusions contained herein are those of the authors and should not be
interpreted as necessarily representing the official policies,
either expressed or implied, of ODNI, IARPA, or the U.S.
Government. The U.S. Government is authorized to reproduce and distribute reprints for governmental purposes not
withstanding any copyright annotation therein.

{\small
\bibliographystyle{ieee}
\bibliography{nikhalriggan_ijcb2023}
}

\end{document}